\documentclass[a4paper, 10pt, conference]{ieeeconf}      

\IEEEoverridecommandlockouts                              

\usepackage{graphicx} 
\usepackage{times} 
\usepackage{amsmath} 
\usepackage{amssymb}  
\usepackage{multirow,color}
\usepackage{cite}
\usepackage{algorithm}
\usepackage{algpseudocode}
\usepackage{varwidth}
\usepackage{subfig}
\usepackage[table]{xcolor}
\usepackage{booktabs}
\usepackage[hyphens]{url}
\usepackage{bm}

\usepackage{color}

\DeclareMathAlphabet{\bdmath}{OML}{cmm}{b}{it}     
\mathversion{normal}

\def\0{\mathbf{0}}

\sloppypar

\title{\LARGE \bf
Analyzing Material Recognition Performance of Thermal Tactile Sensing using a Large Materials Database and a Real Robot
}

\author{Haoping Bai, Haofeng Chen, Elizabeth Healy, Charles C. Kemp, Tapomayukh Bhattacharjee
 \thanks{H. Bai and H. Chen were with Georgia Institute of Technology and E. Healy was with Cornell University when the work was done, C. C. Kemp is with the Healthcare Robotics Lab, Georgia Institute of Technology, and T. Bhattacharjee is with the EmPRISE Lab, Cornell University}
 \thanks{Work was done at Georgia Institute of Technology and Cornell University}
 \thanks{T. Bhattacharjee is the corresponding author
 \{\tt\small tapomayukh@cornell.edu\}.
}%
}

\renewcommand{\baselinestretch}{0.97}

\begin{document}


\maketitle
\thispagestyle{plain}
\pagestyle{plain}


\begin{abstract}
In this paper we focus on analyzing the thermal modality of tactile sensing for material recognition using a large materials database. Many factors affect thermal recognition performance, including sensor noise, the initial temperatures of the sensor and the object, the thermal effusivities of the materials, and the duration of contact. To analyze the influence of these factors on thermal recognition, we used a semi-infinite solid based thermal model to simulate heat-transfer data from all the materials in the CES Edupack Level-1 database. We used support-vector machines (SVMs) to predict $F_1$ scores for binary material recognition for $2346$ material pairs. We also collected data using a real robot equipped with a thermal sensor and analyzed its material recognition performance on $66$ real-world material pairs. Additionally, we analyzed the performance when the models were trained on the simulated data and tested on the real-robot data. Our models predicted the material recognition performance with a $0.980$ $F_1$ score for the simulated data, a $0.994$ $F_1$ score for real-world data with constant initial sensor temperatures, a $0.966$ $F_1$ score for real-world data with varied initial sensor temperatures, and a $0.815$ $F_1$ score for sim-to-real transfer. Finally, we present some guidelines on sensor design and parameter choice for thermal recognition based on the insights gained from these results that would hopefully enable robotics researchers to use this less-explored tactile sensing modality more effectively during physical human-robot and robot-object interactions. We release our simulated and real-robot datasets for further use by the robotics community.

\end{abstract}

\IEEEpeerreviewmaketitle

\newcommand\pkg[1]{\textsf{#1}}
\newcommand\file[1]{\texttt{#1}}

\newcommand{\norm}[1]{\left\lVert#1\right\rVert}


\section{Introduction}\label{sec:intro}
Material recognition using thermal sensing is relatively unexplored in robotics when compared with other haptic sensing modalities such as force sensing. Under some conditions, robots can use this sensing modality to recognize contact with materials and objects that have distinct thermal properties useful for manipulation~\cite{bhattacharjee2021material, bhattacharjeematerial, bhattacharjee2018multimodal, WADE20171, doi:10.1021/acsami.1c17923, bhattacharjee2016data}. For example, a robot might come in contact with a bed frame or a mattress while assisting a person with a disability who is lying down. Recognizing that the object in contact is wood might help a robot infer that it is in contact with the bed frame instead of the human body or the mattress and thus, the robot might alter its actions. This is particularly relevant in physical human-robot and robot-object interaction scenarios where the environments are cluttered such as during robotic caregiving in unstructured homes. These environments may not always have clear line-of-sight and complementary touch sensing modalities such as thermal sensing can be useful. However, the performance of material recognition with thermal tactile sensing varies considerably with different sensor, object, and environment properties. 

\begin{figure}
\centering
\includegraphics[width=0.9\columnwidth]{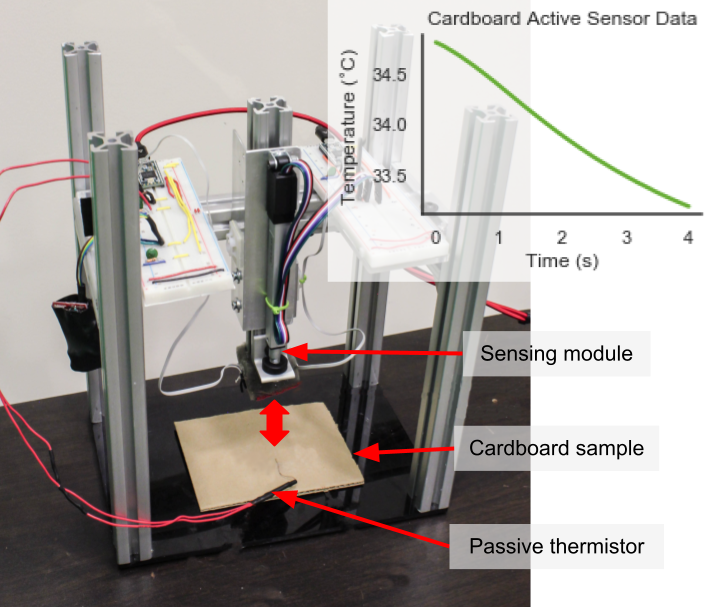}
\caption{\label{fig:1dof}A 1-DoF Robot with an active thermal sensing module reaching to touch a cardboard material sample and the resulting active thermal sensor data.}
\vspace{-7mm}
\end{figure}

\begin{figure*}[!ht]
\centering
\includegraphics[width=0.9\textwidth]{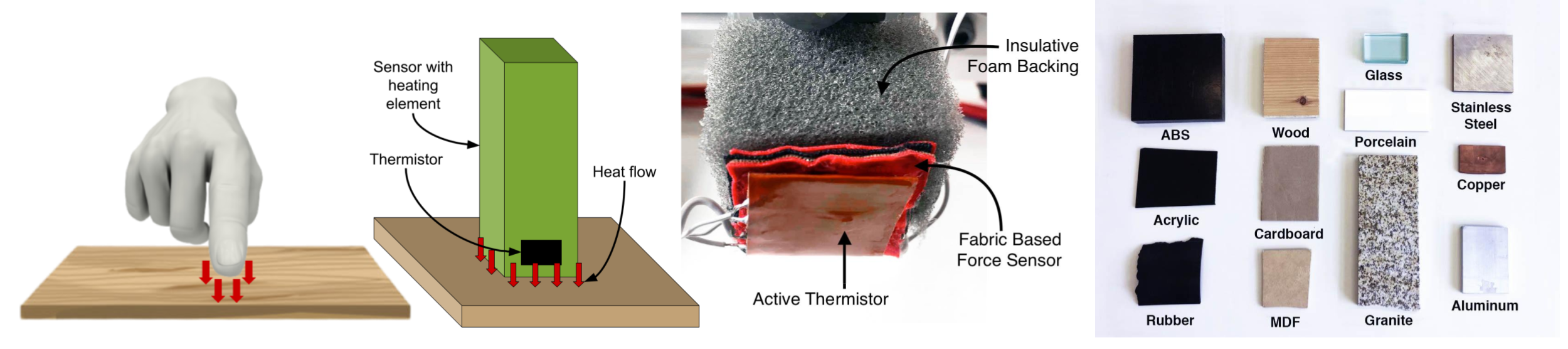}
\caption{\label{fig:temp_flow}Going from left to right: a representation of heat transfer from body to object \cite{finger_image}, a diagram representing our model of the sensor in contact with a material, the sensing module with fabric-based force sensor and an active thermal sensor, the 12 materials for which data was collected with the 1-DoF robot. }
\vspace{-7mm}
\end{figure*}

To gain insight on this variability, we analyzed the material recognition performance on a wide range of materials from a large materials database with different properties of the object, the sensor, the environment, and the contact made between the object and the sensor. We used a physics-based model with a semi-infinite solid assumption for modeling heat-transfer from the heated sensor to the object and added Gaussian i.i.d noise to model the effect of noise. This model can account for the variability in the initial conditions of the sensor and the object, the sensor and object thermal properties, as well as noise. Using this model, we can generate simulated time-series heat-transfer data given sensor and object parameters as well as their initial temperature conditions for a large set of physically-meaningful parameters. We simulated the data for $69$ materials from the publicly-available CES Edupack Level-1 Database~\cite{ashby2008ces}. We used this simulated time-series data to train and evaluate our machine learning model for a total of $2346$ material pairs. We also performed real-world experiments using a real robot equipped with a thermal sensor collecting data from material blocks and comparing the material recognition performance of $66$ material pairs. Additionally, we analyzed how our data-driven model, trained on the simulated data, performed on data collected using the real robot. 


The use of thermal sensing in robotics, though relatively unexplored compared to other modalities of tactile sensing such as force and vibration sensing, is not new. Many researchers used integrated thermal and tactile sensing systems~\cite{engel2005polymer, engel2006flexible, takamuku2008robust, liu2008sensor, mittendorfer2011humanoid,kim_skin2014} for material recognition. Most of these studies, including our own previous work~\cite{bhattacharjee2021material, bhattacharjeematerial, bhattacharjee2018multimodal, WADE20171, doi:10.1021/acsami.1c17923, bhattacharjee2016data}, use thermal sensing for material recognition under specific conditions such as fixed sensor noise, fixed initial conditions, or fixed sensor-object contact duration with a small number of materials. While these provide great insights, they still do not capture real-world variability due to a variety of object, sensor, and environment conditions that thermal sensing is susceptible to. Therefore, it is still unclear as to what benefit this sensing modality provides for the robotics community, when compared to modalities with a extensive body of work such as audition, vision, and force sensing. Specifically in the study of thermal tactile material recognition, some researchers used the SynTouch BioTAC sensor~\cite{biotac_sensor}. Xu et al.~\cite{xu2013tactile} used the BioTAC sensor to measure the temperature derivative and other multimodal sensor data, and used Bayesian exploration and reinforcement learning techniques to identify ten objects with $99\%$ accuracy. Chu et al.~\cite{chu2015robotic} used the BioTAC sensor on PR2 robots to get haptic data. They used HMMs for modeling and used SVMs to assign adjectives to the collected haptic signals automatically. Kerr et al.~\cite{kerr2013material} used the BioTAC sensor on six material groups and used the derivative of the temperature (TAC) and the dynamic thermal conductivity (TDC) data to get $73\%$ accuracy with ANNs.

This paper takes a deep dive into thermal tactile sensing by leveraging a large material database and analyzes the effect of material thermal effusivities, initial temperatures, and noise on the material recognition performance for a wide range of simulated and real-world materials. Our work demonstrates the usefulness of material databases and simulated thermal sensor data in material recognition as well as explores the feasibility of using data-driven methods for sim-to-real transfer. To advance the use of thermal sensing in the robotics community, we release our simulated and real-robot datasets to stimulate further research across the robotics community\cite{harvard_500_data, harvard_level1_data,harvard_12_sim, OAHD}. Finally, we provide some guidelines for thermal sensor design and parameter choice for a desired material recognition performance, given material and sensor properties, as well as environmental conditions.

\section{Physics-based Models}\label{sec:model}
In this paper, we focus on heat-transfer based thermal sensing, which involves a tactile sensor with a heating element and a temperature sensor touching an object. We refer to this as `active' thermal sensing in contrast to `passive' thermal sensing, which we use to refer to a temperature sensor alone making contact with an object. During active thermal sensing, when the tactile sensor, which is heated above room temperature, comes in contact with an object at room temperature, heat transfers away from the sensor into the object. This heat-transfer is dependent on the sensor and object thermal properties, the initial temperature conditions of the sensor and the object, as well as the noise due to various sensor and environmental conditions. A robot can sometimes use the difference in this heat transfer for different materials to distinguish them. Here we present a physics-based model of the heat transfer process between a heated sensor and a material that will output time-series heat-transfer data later used to train and validate our models.

\subsection{Semi-infinite Solid Model}\label{ssec:model_fwd}

We modeled the heat transfer process between a heated thermal sensor and a block of material as heat conduction between two semi-infinite solids \cite{cen_ch4, mathis2000new}. Figure \ref{fig:temp_flow} shows the diagram that represents this model. 

In the model, the initial temperature of the object,  $\mathcal{T}_{o}=T_{o}(t=0)$, is equal to the ambient temperature, $\mathcal{T}_{a}$ and we set the initial sensor temperature, $\mathcal{T}_{s}=T_{s}(t=0)$, higher than $\mathcal{T}_{a}$. The contact surface at $x=0$, where $x$ is the distance from the thermistor to the surface, has a temperature $\mathcal{T}_{c}$ that remains constant and is given by
\begin{align}\label{eq01}
{\mathcal{T}_{c}} &= \frac{{\left( {{\mathcal{T}_{s}} e_{s} + {\mathcal{T}_{o}} e_{o}} \right)}}{{\left( {e_{s} + e_{o}} \right)}} \quad with & e_{s} = \frac{{{k_{s}}}}{{\sqrt {{\alpha _{s}}} }}, e_{o} = \frac{{{k_{o}}}}{{\sqrt {{\alpha _{o}}} }}
\end{align}
where  ${\alpha _{o}}$ and $k_{o}$ are the coefficients of thermal diffusivity and thermal conductivity of the object respectively, and ${\alpha _{s}}$ and $k_{s}$ are the coefficients of thermal diffusivity and thermal conductivity of the sensor respectively. Given  $\mathcal{T}_{s}$ and $\mathcal{T}_{c}$, we can find the temperature of the sensor at any time, $t\geq0$. 
\begin{align}\label{eq02}
\begin{split}
\begin{gathered}
  {T_{s}}\left( {x,t} \right) = {\mathcal{T}_{s}} + \left( {{\mathcal{T}_{c}} - {\mathcal{T}_{s}}} \right) * erfc\left( {\frac{x}
{{2\sqrt {{\alpha _{s}}t} }}} \right)\\ 
\end{gathered}
\end{split}
\end{align}
where $erfc$ is the complimentary error function given by
\begin{align}\label{eq03}
\begin{split}
erfc(z)= \frac{2}{ \sqrt{\pi} }  \int_z^ \infty  e^{-r^2} dr
\end{split}
\end{align}



\begin{figure}
\vskip\baselineskip
\centering
\includegraphics[width=1\columnwidth]{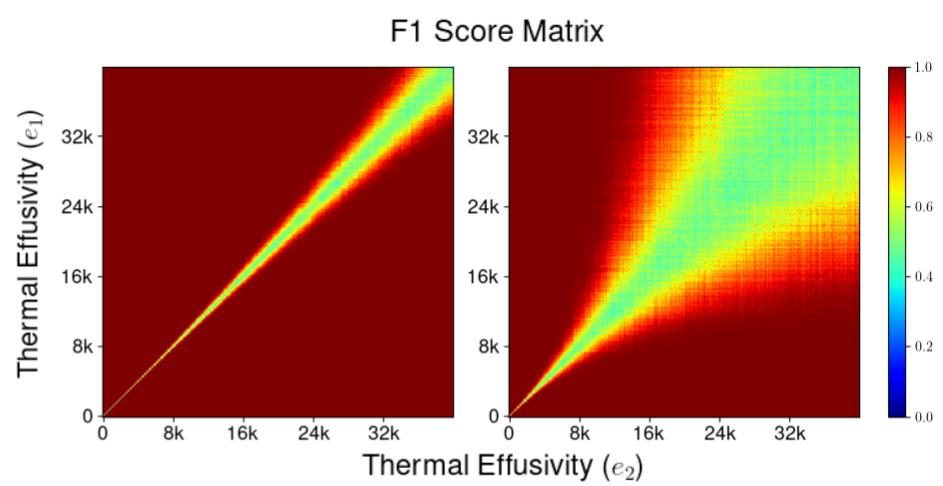}
\caption{\label{fig:f1mat_0_50}$F_1$ score matrices for our SVM model with $t_{contact}$ = $4.00s$ (left) and $t_{contact}$ = $1.00s$ (right) ($\mathcal{T}_{s}$ = $35^{\circ}$C, $\sigma$ = $0.05$, $\mathcal{T}_{a}$ = $25^{\circ}$C). This demonstrates that increased contact duration results in increased performance and that rising $\delta(e)$ makes it more difficult to distinguish large effusivities.}
\vspace{-7mm}
\end{figure}

\subsection{Noise Model}\label{ssec:sensor_model}
Note that during each temperature measurement, the measurement of the sensor also includes noise and other sources of uncertainty. To account for this, we introduce an additive Gaussian noise, $Z_{i}$, with zero mean and variance $\sigma^2$ to each temperature measurement. The underlying assumption is that the deviation of each sensor reading from the actual sensor temperature caused by the uncertainty due to various conditions can be modeled as an independent normal random variable.

With noise taken into consideration, the final sensor model is given by
\vspace{0mm}
\begin{align}\label{eq04}
\begin{split}
\begin{gathered}
  {T_{sens}}\left( {x,t} \right) = {\mathcal{T}_{s}} + \left( {{\mathcal{T}_{c}} - {\mathcal{T}_{s}}} \right) * erfc\left( {\frac{x}
{{2\sqrt {{\alpha _{s}}t} }}} \right) \\ + Z \sim \mathcal{N}\left (0, \sigma^2 \right )\\ 
\end{gathered}
\end{split}
\end{align}

This modified model can help us analyze the effect of noise on the performance of material recognition.

\section{Experiment Methods}\label{eval}
To gain insight into material recognition performance with thermal sensing data under varying conditions, we used the physics-based model to simulate data as well as collected real robot data. We selected support-vector machines (SVMs)\footnote{We also experimented with other models such as Gaussian Naive Bayes (GNB) and linear discriminant analysis (LDA) but found SVMs to be the most robust across conditions and data sources while providing consistent results}, as our data-driven method because it is a simple method that is widely used~\cite{8594469, KERR201894, app9122537} and does not require a lot of data, which was crucial for us given the constraints of collecting real-world physical interaction data. Also, we have previously achieved success with SVMs for material recognition tasks using active thermal sensing \cite{bhattacharjee2016data, bhattacharjeematerial, bhattacharjee2018multimodal}. We used the implementation of binary SVM provided by the scikit-learn package \cite{scikit-learn} in Python with a linear kernel. To produce feature vectors for training, we used both raw temperature and estimated local slope from each trial of experiment, and concatenated them into a single feature vector. Using SVMs, we performed a four-part evaluation in which we investigate what factors influence performance, how they influence performance, and whether simulated data is a viable option for training.

\begin{itemize}

\item First, we focus on classifying simulated sensor time-series data for any two different arbitrary thermal effusivities. We use the entire range of physically feasible thermal effusivities to compare the performance and analyze the effect of noise and sensor initial condition on the performance. 

\item Second, we focus on the prediction of the models in binary material recognition for all materials in the CES-Edupack Level 1 Database \cite{ashby2008ces} using simulated time-series heat-transfer data resulting in $2346$ material pair comparisons. We simulate data using consistent sensor initial conditions.

\item Third, we focus on binary material recognition of $12$ real-world materials. We analyze the prediction of the model in binary material recognition tasks for real-world time-series heat-transfer data collected using a 1-DoF robot from $12$ different materials under both consistent and varied sensor initial conditions, resulting in $66$ real-world material pair comparisons for each condition type.

\item Finally, we focus on material recognition performance on the real-world data from the 1-DoF robot when models were trained on only simulated data for the same $12$ materials. 

\end{itemize}

We used $F_1$ scores as a metric of performance for all the four cases and also calculated the number of indistinguishable material pairs for each case to provide more insight into the material recognition performance.


\section{Evaluation: Different Thermal Effusivities}\label{part_1}
In this first set of experiments, we obtain the $F_1$ scores of the model, trained on simulated time-series data, for classifying any two different arbitrary thermal effusivities. Given a reference thermal effusivity value, we are interested in the minimum effusivity difference $\delta(e)$ required to obtain a binary classification $F_1$ score greater than or equal to a desired performance ($\Phi$). In this paper, we set a threshold of $\Phi = 0.9$. This means we consider any effusivity pair with $F_1 \geq 0.9$ classification score as distinguishable. We require a high performance metric as we want to be confident that the two materials are distinct. 



When a thermal sensor comes in contact with a material, the heat-transfer data is affected by sensor noise, initial sensor temperature as well as the contact duration. Therefore, we analyze the effect of these parameters on $F_1$ score by varying the quantities as given below:

\begin{itemize}
\item Noise $Z \sim \mathcal{N}\left (0, \sigma^2 \right )$ : $\sigma=0.01$ , $\sigma=0.05$ and $\sigma=0.1$
\item Initial Sensor Temperature $\mathcal{T}_{s}$ : $30^{\circ}$C and $35^{\circ}$C 
\item Contact Duration $t_{contact}$: $1.00$s, $2.00$s, $3.00$s, and $4.00$s
\end{itemize}

We estimated the minimum distinguishable difference $\delta(e)$ for every effusivity value $e$ for the above conditions. We generated noisy data using the physics-based model and performed a 3-fold cross-validation over each unique effusivity value pair and reported the $F_1$ score.

\begin{figure}[t!]
\centering
\vskip\baselineskip
\centering
\includegraphics[width=1\columnwidth]{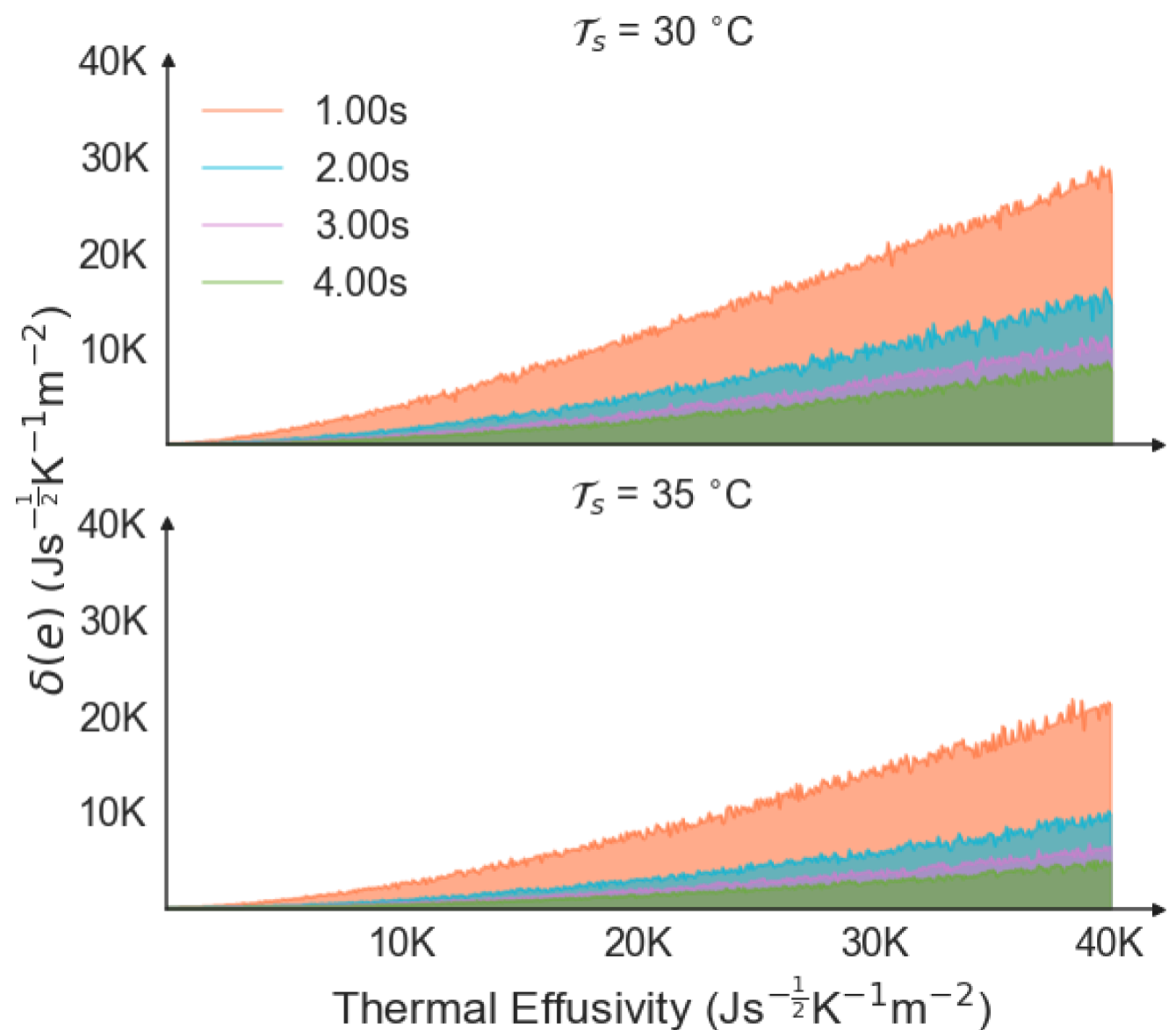}
\caption{Effect of Initial Condition on $\delta(e)$ with fixed noise $\sigma = 0.05$: $\mathcal{T}_{s}$ = $30^\circ$C (Top), $\mathcal{T}_{s}$ = $35^\circ$C (Bottom) ($\mathcal{T}_{a}$ = $25^\circ$C). We see that $\delta(e)$ is lower with longer contact duration and larger temperature difference between sensor and material.}\label{fig:theory-init-cond}
\vspace{-7mm}
\end{figure}

\begin{figure*}[ht!]
\centering
\vskip\baselineskip
 \centering
 \hspace{0.5mm}
\includegraphics[width=1\textwidth]{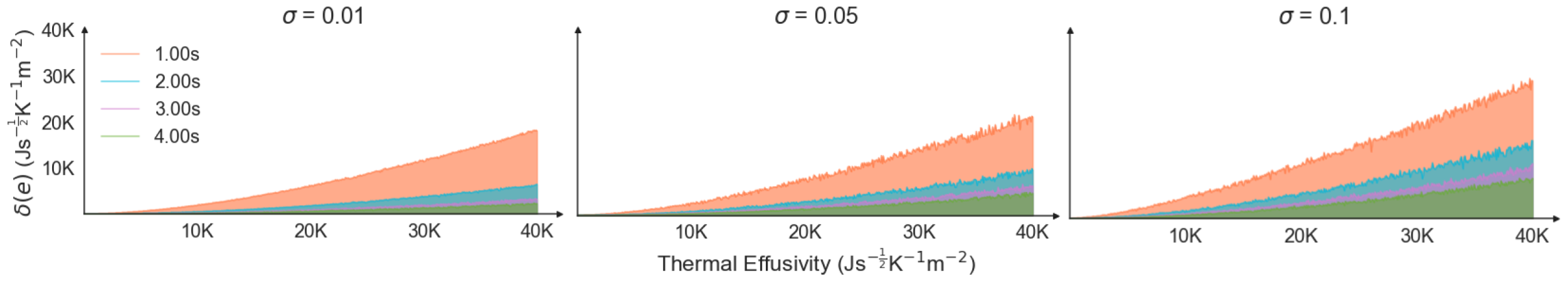}
\caption{Effect of Noise on $\delta(e)$ with fixed initial condition $\mathcal{T}_{s} = 35^\circ$C and $\mathcal{T}_{a} = 25^\circ$C: $\sigma=0.01$ (Left), $\sigma=0.05$ (Middle), $\sigma=0.1$ (Right). This demonstrates that lower noise produces lower $\delta(e)$ values.}\label{fig:theory-noise}
\vspace{-7mm}
\end{figure*}

\subsection{Data Collection}\label{ssec:sensor_data}
In order to account for a sufficiently large thermal effusivity range, we referred to the CES EduPack 2016\cite{ces_edupack} Level 1 material database. Of all the included materials, Rigid Polymer Foam (LD) has the minimum effusivity value of $3.05 \times 10^1$ Js$^{-\frac{1}{2}}$K$^{-1}$m$^{-2}$ and Copper Alloy has the maximum effusivity value of $3.68 \times 10^4$ Js$^{-\frac{1}{2}}$K$^{-1}$m$^{-2}$. Therefore, we sampled effusivity values in the range $(0, 4.00 \times 10^4] \,\,$ Js$^{-\frac{1}{2}}$K$^{-1}$m$^{-2}$. We discretized the range to $500$ equal intervals resulting in $124,750$ effusivity pairs. We can think of each interval as a material category, and an instance of the material category can take on any effusivity value within the interval. 


Given an effusivity value $e$, we constructed the time series heat-transfer data based on the semi-infinite solid model defined in \ref{ssec:model_fwd}. We use $e_{s} = 892$ (Js$^{-\frac{1}{2}}$K$^{-1}$m$^{-2}$), and $\alpha_{s} = 1.19 \times 10^{-9}$ (m$^2$s$^{-1}$) similar to our real-world sensor parameters. We set $\mathcal{T}_{a}$ to $25^\circ$C and set the sampling rate to be $200$ Hz, similar to our real-world sensor sampling rate. 

We generated $100$ trials for each effusivity interval by uniformly sampling from the effusivity interval and generating simulated data with the sampled effusivity. 



\subsection{Results and Discussion}\label{ssec:sim_result}

In this section, we present the results of the above evaluation. Figure \ref{fig:f1mat_0_50} shows two $F_1$ score matricies with pairwise $F_1$ scores for all effusivity values using our SVMs. We obtained each matrix using $\mathcal{T}_{s}$ = $35^{\circ}$C and $\sigma=0.05$ and varying $t_{contact}$ from $4.00$s to $1.00$s. Table \ref{percent_edges_avg_effu} shows the $F_1$ scores and the percentage of indistinguishable effusivity combinations calculated based on the $F_1$ score matrices with $\Phi = 0.9$ for a single contact duration of $t_{contact}=2.00s$. 



\begin{table}[]
\centering
\caption{Material recognition performance on simulated data using SVMs with $t_{contact} = 2.00$s}
\label{percent_edges_avg_effu}
\resizebox{\columnwidth}{!}{%
\begin{tabular}{@{}c|ccc|cc@{}}
\midrule
\multicolumn{2}{c}{} & \multicolumn{2}{c}{\textbf{Simulated Effusivities}} & \multicolumn{2}{c}{\textbf{Simulated Materials}} \\
 \hline
 & \multicolumn{1}{l}{}               & \multicolumn{2}{|c|}{\textit{Temperature Difference}} & \multicolumn{2}{c}{\textit{Temperature Difference}} \\
Indistin & \multicolumn{1}{l|}{\textit{Noise}} & \textit{5°C}     & \textit{10°C}                    & \textit{5°C}             & \textit{10°C}            \\ \cline{2-6}
guishable & \multicolumn{1}{c|}{0.1}           & 20.25\%          & \multicolumn{1}{c|}{16.54\%}     & 47.76\%                  & 31.56\%                  \\
Pairs & \multicolumn{1}{c|}{0.5}           & 16.41\%          & \multicolumn{1}{c|}{14.96\%}     & 30.87\%                  & 18.28\%                  \\
 & \multicolumn{1}{c|}{0.01}          & 13.51\%          & \multicolumn{1}{c|}{13.04\%}     & 20.19\%                  & 11.28\%                  \\ 
\midrule
 & \multicolumn{1}{l}{}               & \multicolumn{2}{|c|}{\textit{Temperature Difference}} & \multicolumn{2}{c}{\textit{Temperature Difference}} \\
 & \multicolumn{1}{l|}{\textit{Noise}} & \textit{5°C}     & \textit{10°C}                    & \textit{5°C}             & \textit{10°C}            \\ \cline{2-6}
$F_1$ & \multicolumn{1}{c|}{0.1}           & 0.815           & \multicolumn{1}{c|}{0.882}      & 0.931                   & 0.943                   \\
Scores & \multicolumn{1}{c|}{0.5}           & 0.885           & \multicolumn{1}{c|}{0.934}      & 0.945                   & 0.952                   \\
 & \multicolumn{1}{c|}{0.01}          & 0.900           & \multicolumn{1}{c|}{0.945}      & 0.958                   & 0.960                   \\ 
\bottomrule
\end{tabular}%
}
\vspace{-4mm}
\end{table}

\subsubsection{Effect of Contact Duration}\label{ssec:sim_contact_dur}\

To analyze the effect of contact duration on classification
performance, we truncated the time series data at different time lengths
. Figure \ref{fig:theory-init-cond} shows the minimum distinguishable difference $\delta (e)$ curves calculated based on the SVM results. As expected, in each plot, with increased length of contact duration, the expected material recognition performance improves. 
(See Section \ref{sec:discussion}).

\subsubsection{Effect of Initial Condition}\label{ssec:sim_init_cond}\
Figure \ref{fig:theory-init-cond} shows the results from our SVMs for both $\mathcal{T}_{s}=30^\circ$C and $\mathcal{T}_{s}=35^\circ$C initial conditions. 
By comparing the $\mathcal{T}_{s}=30^\circ$C graphs with the $\mathcal{T}_{s}=35^\circ$C graphs in Fig.\ref{fig:theory-init-cond}, we observe that larger initial temperature difference ($\mathcal{T}_{s}=35^{\circ}C$) between sensor and ambient environment produces a lower $\delta(e)$ curve. In other words, our SVMs predict that a larger initial temperature difference between sensor and measured object can help in material recognition, as it generates more distinguishable heat transfer data for materials (See Section \ref{sec:discussion}).


\subsubsection{Effect of Noise}\label{ssec:sim_noise}
Figure \ref{fig:theory-noise} shows the results of our SVMs for different levels of noise. 
By comparing the three plots ($\sigma=0.01$, $\sigma=0.05$, $\sigma=0.1$ left to right), 
in Fig. \ref{fig:theory-noise}, we observe that simulations with a noise level $\sigma=0.1$ produce the highest $\delta(e)$ values. Again, our models predict that thermal sensors with lower noise help in material recognition (See Section \ref{sec:discussion}).

\section{Evaluation: Materials Database}

In this set of experiments, we mapped the previous results obtained using different thermal effusivity values to actual material effusivity values. 
We obtained thermal effusivity values of all $69$ materials from CES EduPack Level 1 database \cite{ashby2008ces}. Figure \ref{fig:effu_dist} shows the effusivity ranges of these materials. We looked up binary material classification results for all possible pairs of effusivity values corresponding to $69$ materials ($2346$ pairs) from our previous results in Section \ref{ssec:sim_result} to find out what materials are distinguishable with $F_1$ score greater than $0.9$.

\subsection{Node Graphs of Material Pairs}\label{ssec:ideal_indis}\
To visualize whether any two materials from the CES EduPack Level 1 database \cite{ashby2008ces} are distinguishable, we generated a node-graph based on their $F_1$ scores where each node represents a material. The node-graph has the following characteristics: 
\begin{itemize}
\item An edge between two material nodes represent that they are indistinguishable. Note $\Phi=0.9$.
\item The radius of a material node is proportional to its thermal effusivity. 
\item CES Edupack divides all materials into four large categories such as metals / alloys, ceramics / glasses, polymers / elastomers, and composites / foams / natural. A material node's color signifies which category the material belongs to.
\item The thickness of the edge connecting two materials is inversely proportional to their $F_1$ score. This means that the thicker the edge, the more difficult it is to distinguish the material nodes.
\item The relative position of the nodes has no relation with any physical property.
\end{itemize}

Note, each material in the CES Edupack database \cite{ashby2008ces} has a range of thermal effusivity values that it can have. To find the average $F_1$ score for gold and silver for example, we find the average of $F_1$ scores for the binary classification between all possible combinations of gold effusivities and silver effusivities. In our case, the average $F_1$ score can be calculated based on the $F_1$ score matrix, as shown in Fig. \ref{fig:f1mat_0_50}.

\begin{figure}[!t]
\centering
\vskip\baselineskip
\centering
\includegraphics[width=1\columnwidth]{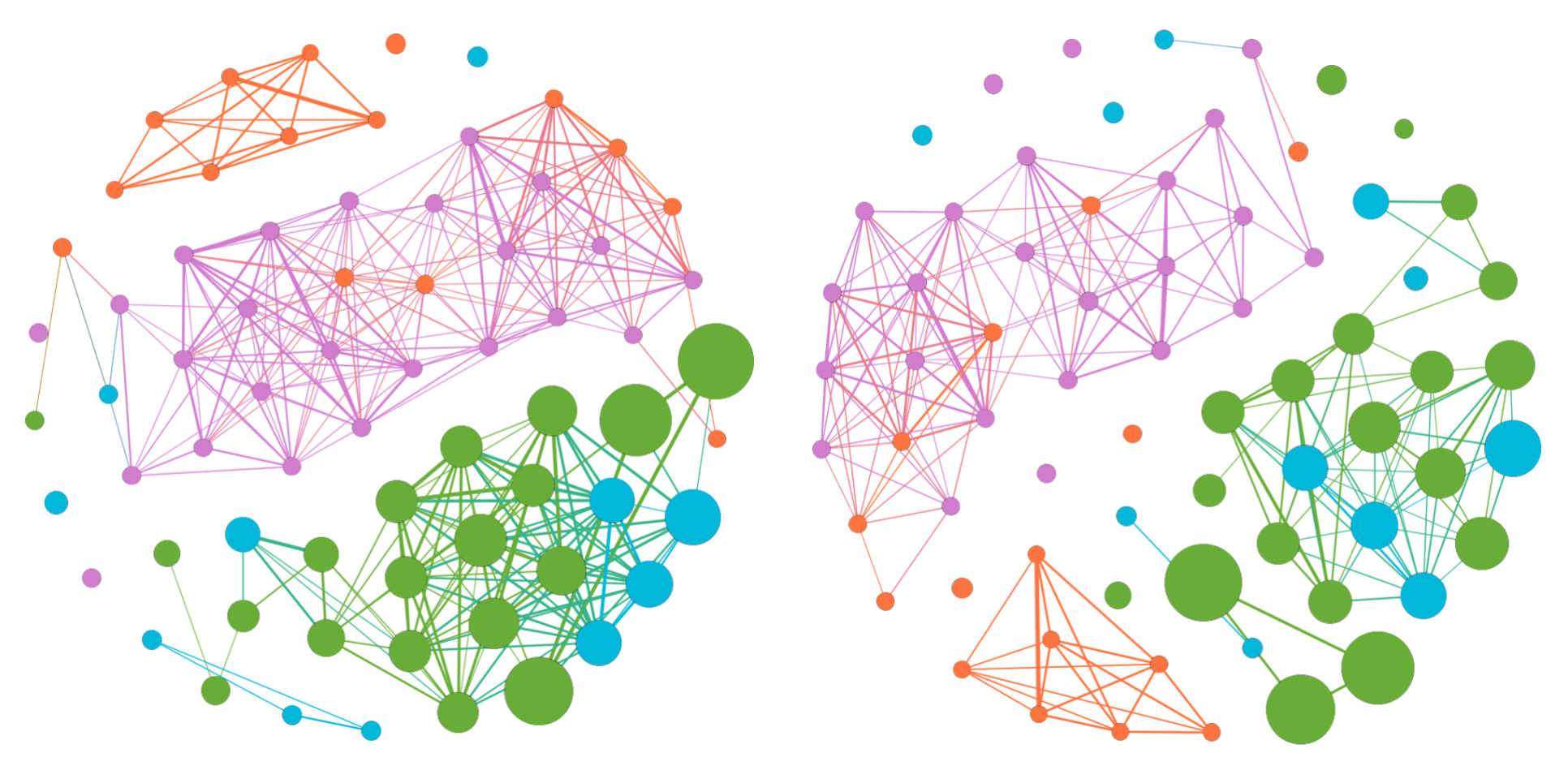}


\includegraphics[height=1cm]{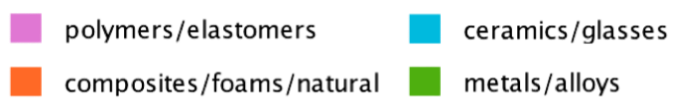}
\caption{Node Graphs of Material Pairs with $\mathcal{T}_{s}=30^\circ$C (left) and $\mathcal{T}_{s}=35^\circ$C (right) ($\sigma=0.05$ and $\mathcal{T}_{a}=25^\circ$C. This further shows that it is easier to distinguish between materials with a larger temperature difference between sensor and material. }\label{fig:graph_of_indis_pairs}
\vspace{-7mm}
\end{figure}

Figure \ref{fig:graph_of_indis_pairs} shows the results of $\mathcal{T}_{s}=30^{\circ}$C and $\mathcal{T}_{s}=35^{\circ}$C with $t_{contact}=2.00$s and $\sigma=0.05$ noise. From the figures, we can again see how initial temperature difference between surface and sensor affects distinguishibility as the graph with a higher $\mathcal{T}_{s}$ has fewer connected nodes and thus fewer indistinguishable pairs. Additionally, we note that there are three to four connected components in the node-graph and these connected components tend to have a majority of the material nodes in a particular category.
This further means that a material belonging to one of these categories has a higher probability of being distinguished from a material in another category than in its own category. 
We can also see some densely connected components in the graph. For example, metals are densely connected together, which agrees with our observation in Fig. \ref{fig:f1mat_0_50}, as rising $\delta (e)$ makes it harder to distinguish two materials with larger effusivities.

\begin{figure}[!t]
\centering
\includegraphics[width=1\columnwidth]{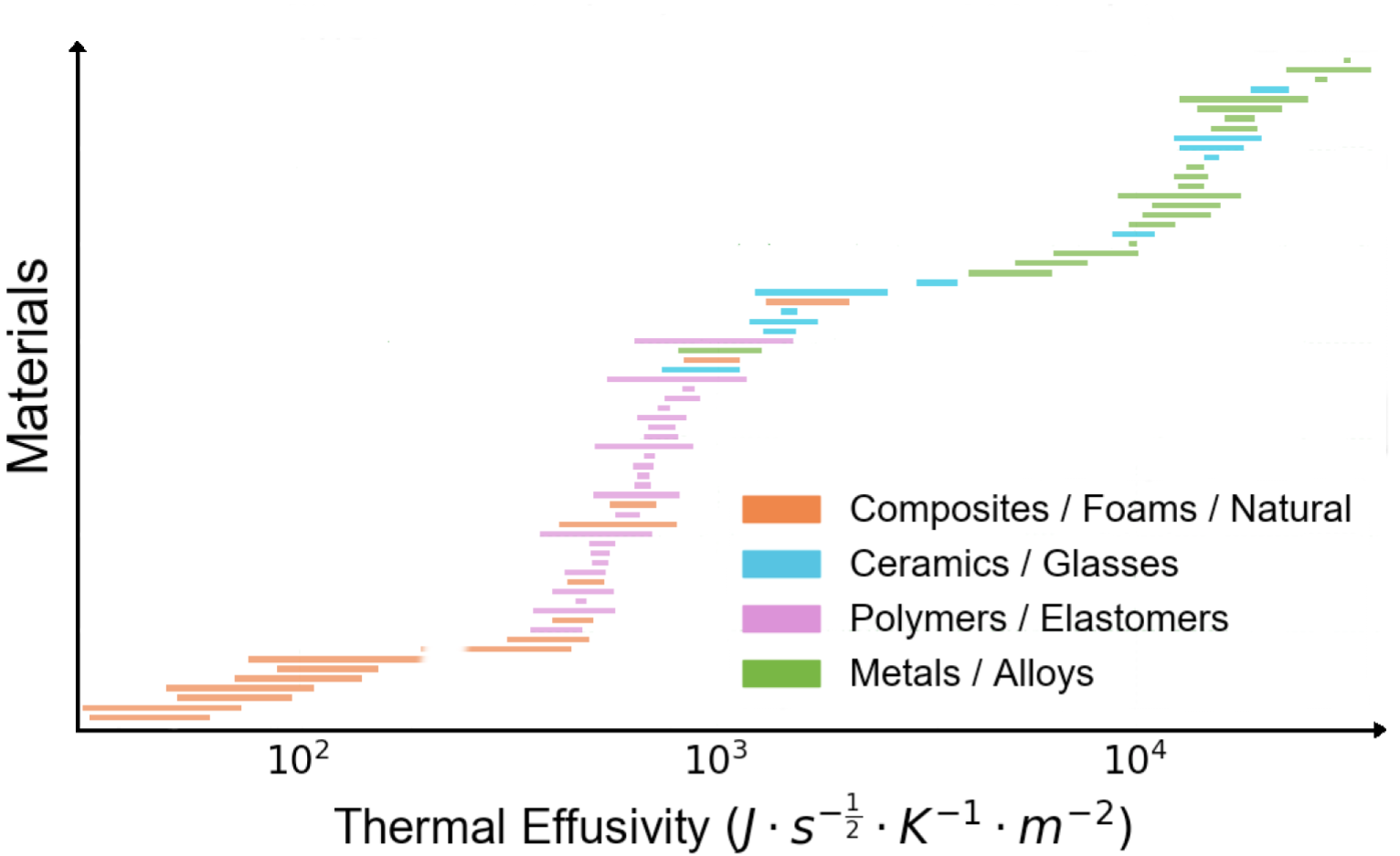}
\caption{\label{fig:effu_dist}Effusivity Distribution of the $69$ Materials in CES Edupack Level 1 \cite{ashby2008ces} in Logarithmic Scale}
\vspace{-7mm}
\end{figure}

The observed connected components also agree well when compared with the effusivity ranges provided in Fig.\ref{fig:effu_dist}. Metals, with large effusivity values, are generally difficult to distinguish amongst themselves because their effusivity values are so large that they dominate $\mathcal{T}_{c}$ (Eq.\ref{eq01}) to a value very close to the ambient temperature, rendering the $T_{s}$ curves indistinguishable.

Polymers / elastomers are more densely connected than the metals. Looking at Fig. \ref{fig:effu_dist}, we see that the effusivity ranges of this group are very similar, thus making them harder to distinguish.

As shown in Table. \ref{percent_edges_avg_effu}, the number of edges present in the graph is consistent with the observation we made in Section \ref{ssec:sim_result}, that a larger initial temperature difference between sensor and material and less noise leads to more distinguishable material pairs. 


\section{Evaluation: Real Robot Data}\label{sec:approach}

\subsection{Experimental Setup}\label{ssec:setup}
The 1-DoF robot consists of a linear actuator, two Teensy 3.2 microcontrollers, a passive sensing thermistor, and an active sensing module. The active sensing module consists of the Thorlabs HT10K Flexible Polyimide Foil Heater with 10 kOhm Thermistor \cite{thorlabs} (heating element and a temperature sensor) on a fabric based force sensor \cite{bhattacharjee2013tactile} which is backed by thermal insulation foam. The passive sensing thermistor uses the fast-response $10 k \Omega$ NTC thermistor (EPCOS B57541G1103F) \cite{epcos}.


The materials used for this set of experiments are shown in Fig. \ref{fig:temp_flow}. We selected these materials in order to have uniform representation of materials from all  four categories (metals, ceramics, polymers, and composites) from the CES Edupack database \cite{ashby2008ces}. 
We selected the $12$ materials such that there are distinguishable and indistinguishable material pairs. We estimated this by using the mid-point of the effusivity range of these materials.


\subsection{Experimental Procedure}\label{ssec:procedure}
We used a Python script on a separate Dell Optiplex 9010 Computer equipped with Intel(R) Core(TM) i7-3770 CPU at 3.40 GHz running 32-bit Ubuntu 12.04.2 LTS system with Linux Ubuntu 3.5.0-54-generic kernel to control the device through a serial link with the Teensy 3.2 microcontrollers. Before reaching down and contacting the sample, the device waits at $15$ mm above the sample, to allow a voltage supply to generate heat based on an integral controller such that the active sensing thermistor maintains a desired temperature. Upon contact with the material, the integral controller stops so as not to interfere with the natural heat-transfer from the sensor to the material. The micro-controllers record the active sensing thermistor and the passive sensing thermistor readings at $200$ Hz for $10$ seconds. Note, the sensor has an insulated foam backing which makes the sensor compliant and thus, to ensure that there is complete contact between the material and the sensor's flat surface, we use a force threshold of $5$N to detect the onset of contact. Also, we do not use the passive thermistor data for any material recognition purposes. The robot then raises the sensing module and waits for $20$ seconds before starting the next trial. Using the FLIR Tau 2 324 $7.5$mm Thermal Imaging Camera Core (46324007H-FRNLX), we found that $20$ seconds was enough for the materials to come back to their initial state. This is to ensure that the material is at a consistent initial condition before the robot touches it at any trial. 

We performed two sets of experiments with the real robot. The first set consisted of $10$ trials each with fixed initial sensor temperature conditions for each material. The second set consisted of $50$ trials each with uniformly varied initial sensor temperature conditions for each material. We uniformly varied the initial sensor conditions between $\mathcal{T}_{s}=30^\circ$C to $\mathcal{T}_{s}=35^\circ$C. We identified the sensor and material parameters as outlined in Appendix \ref{ssec:find_sensor_params}. We performed this set of experiments to simulate contact situations when a robot incidentally touches objects in its environment without the opportunity to adjust its initial sensor conditions. This is a common scenario in manipulation in cluttered and unstructured environments or in assistive scenarios working in close contact with a human body \cite{bhattacharjee2014inferring}.

\subsection{Results}\label{ssec:exp_result}
Trained and tested on real robot data with fixed initial conditions, the SVMs achieved an average $F_1$ score of $0.994$ for binary material recognition across all $66$ material pair comparisons. When trained and tested with varied initial sensor conditions, the models achieved an average $F_1$ score of $0.966$ for across all $66$ pairs.



\begin{table}[]
\centering
\caption{Material recognition performance on real data using SVMs with $t_{contact} = 4.00$s}
\label{real_and_sim_real_results}
\resizebox{\columnwidth}{!}{%
\begin{tabular}{@{}lcccc@{}}
\toprule
                                     & \multicolumn{2}{c}{\textbf{Train Real-Test Real}}     & \multicolumn{2}{c}{\textbf{Train Simulated-Test Real}} \\ \midrule
\multicolumn{1}{l|}{}            & \multicolumn{4}{c}{\textit{Initial Conditions}} \\
\cline{2-5}
\multicolumn{1}{c|}{\textit{Metric}} & \textit{Fixed} & \multicolumn{1}{c|}{\textit{Varied}} & \textit{Fixed}            & \textit{Varied}            \\ \midrule
\multicolumn{1}{c|}{\textit{F1 Score}} & 0.994 & \multicolumn{1}{c|}{0.966} & -- & 0.815 \\ \bottomrule
\end{tabular}%
}
\vspace{-4mm}
\end{table}

\section{Evaluation: Sim-to-Real Transfer}
\label{sec:train_sim_test_real}

To evaluate whether the simulated data can be used to prepare real models and eliminate the lengthy training data collection process, we produced data for each of the $12$ materials using the physics-based model, trained our material recognition data-driven model on that data, and tested the model using the varied initial conditions data collected from the 1-DoF robot.

For the data simulation, in order to produce varied conditions similar to those of the data collected from the robot, we pulled initial sensor temperatures $\mathcal{T}_{s}$ and initial ambient temperatures $\mathcal{T}_{a}$ from truncated normal distributions created using the means and standard deviations of the initial conditions of the real robot data for each material such that
\begin{align}\label{eq04}
\begin{split}
\begin{gathered}
\mathcal{T}_{s} = X \sim \big(min(\mathcal{T}_{sR}) < \mathcal{N}(\mu_{\mathcal{T}_{sR}}, \sigma_{\mathcal{T}_{sR}}) < max(\mathcal{T}_{sR})\big) \\
\mathcal{T}_{a} = X \sim \big( min(\mathcal{T}_{aR}) < \mathcal{N}(\mu_{\mathcal{T}_{aR}}, \sigma_{\mathcal{T}_{aR}}) < max(\mathcal{T}_{aR})\big)
\end{gathered}
\end{split}
\end{align}
where $\mathcal{T}_{sR}$ and $\mathcal{T}_{aR}$ refer to the real-world initial sensor and ambient temperatures.
We used a constant noise of $\sigma=0.075$ after evaluating model test performance with the following noises [$0.01$, $0.025$, $0.05$, $0.075$, $0.1$] and choosing the $\sigma$ that produced the best performance. Additionally, we used the thermal effusitivity values identified for the experiment materials in Table \ref{eff_table} and the values of $e_{s}$ and $\alpha_{s}$ described in Section \ref{ssec:sensor_data}.   

The SVM models, trained on the simulated data and tested on the real robot experiment data, achieved an average $F_1$ score of $0.815$ and found $48.48$\% of the real material pairs indistinguishable. 

Upon closer analysis, the SVMs struggled on material pairs whose thermal effusivities were closer in value or are from the same material category. For example, one model tested on cardboard and wood, which have the smallest thermal effusivities of the 12 materials, struggled with an average $F_1$ of $0.246$ as their difference is smaller than the minimum distinguishable difference $\delta(e)$. Additionally, metals like stainless steel and aluminum, which have very large thermal effusivities, also had a lower average $F_1$ score of $0.233$. As shown in Fig. \ref{fig:f1mat_0_50}, rising $\delta(e)$ makes it harder to distinguish materials with larger effusivities.

\section{Discussion}\label{sec:discussion}

\subsection{Model Limitations and Potential Extensions}
Note the physics-based model which generated the time-series data is based on a semi-infinite solid model assumption, which assumes heat transfer from the active thermal sensor to the material is in one direction only. This assumption is generally valid for a short duration which is characterized by the Fourier Number of the material \cite{ho2007development, yang2008use}. Additionally, the thermal properties of a material change with temperature which we did not account for in our physics-based model. There also exist some thermally ambiguous conditions that make it difficult to distinguish between materials no matter what their effusivities are. Bhattacharjee et al.~\cite{bhattacharjeematerial} found that robots can overcome this ambiguity using two temperature sensors with different temperatures prior to contact. Lastly, it remains to be seen how the performance of machine learning models for binary classification in this paper extend to multi-class classification scenarios.

\subsection{Impact of Contact Area}
Our semi-infinite solid based physics model does not explicitly model contact area, but heat-transfer depends on contact area. In this paper, we used a ‘flat-area’ thermistor similar to one used in \cite{bhattacharjeematerial}. When performing these evaluations with a different ‘point’ sensor (a thermistor of small cross-sectional area) used in \cite{wade2016force}, the SVM’s ability to distinguish between materials with varied initial conditions dropped from an average $96.69$\% to $33.33$\%. 
Depending upon the sensor, application of a larger force might result in a better contact area and contact between two flat surfaces may result in more prominent heat-transfer than contact between a flat surface and a spherical surface (`point' sensor). Also, the `point' sensor parameters may be more susceptible to temperature changes i.e. the thermal effusivity and diffusivity of the `point' sensor may have changed significantly with temperature changes in the sensor. Accounting for the sensor parameter dependence on temperature, the effect of contact area, as well as the force applied during physical contact are interesting directions of future exploration.

\subsection{Guidelines and Implications on Sensor Design}
The combination of machine learning models and the time series data generated from the physics-based model in this paper could be used to design thermal sensors to achieve a desired level of performance and to provide various experimental design guidelines based on their predictions. For example, to be able to distinguish between two materials with thermal effusivities of around $35$k
(Js$^{-\frac{1}{2}}$K$^{-1}$m$^{-2}$) and $20$k (Js$^{-\frac{1}{2}}$K$^{-1}$m$^{-2}$) 
, our results suggest that the robot with the thermal tactile sensor needs to be in contact with the material samples for at least $2$ seconds. This is for a robot with a thermal sensor with $0.05^\circ$C noise and initial temperature $10^\circ$C higher than the material's initial temperature (See Figs. \ref{fig:theory-init-cond} and \ref{fig:theory-noise}). Additionally, this work exemplifies how a materials database can be used not only to explore key factors relevant to material recognition via heat transfer, but also how it can be utilized to simulate data used to train models that perform on real robots.

\appendices

\section{Finding Sensor and Material Parameters}\label{ssec:find_sensor_params}


To identify sensor parameter values (sensor effusivity $e_{s}$ and sensor diffusivity $\alpha_{s}$), we collected $10$ trials of data with fixed initial conditions from each of the $12$ materials (not used for material recognition experiments). We identified the sensor parameter values based on the sum of squared error between experiment temperature data and the ideal temperature data based on the semi-infinite solid model defined in section \ref{ssec:model_fwd}. For each material, we used the Limited-memory BFGS with boundary constraints (L-BFGS-B) \cite{jones2001scipy} algorithm to find its optimal effusivity value, with the boundary constraints given by the thermal effusivity values of materials in the CES EduPack database \cite{ashby2008ces}. In addition, due to noise, it is possible that the heat-transfer started slightly before or after the estimated onset of contact. Thus, we also included a time offset from the onset of contact as an optimization parameter. We used the L-BFGS-B algorithm to find the time offset of the experiment data, and it turned out that the heat transfer started about 0.5s before the estimated onset of contact. We identified the sensor effusivity as $e_{s} = 892$ (Js$^{-\frac{1}{2}}$K$^{-1}$m$^{-2}$), and sensor diffusivity as $\alpha_{s} = 1.19 \times 10^{-9}$ (m$^2$s$^{-1}$). Table \ref{eff_table} shows the identified effusivity values of all materials in this experiment.

\begin{table}[ht!]
\begin{center}
\caption{Thermal Effusivity Values of Materials in the Experiment (Js$^{-\frac{1}{2}}$K$^{-1}$m$^{-2}$)}
\label{eff_table}
\begin{tabular}{@{}cccc@{}}
\toprule
\multirow{2}{*}{\textit{Material}} & \textbf{Thermal effusivity} & \textbf{Min. thermal} & \textbf{Max. thermal} \\
                                     & \textbf{identified} & \textbf{effusivity} & \textbf{effusivity} \\ \midrule
\multicolumn{1}{c|}{Cardboard}       & 336.90   & 196.67   & 452.23   \\
\multicolumn{1}{c|}{Wood}            & 400.95   & 331.00   & 506.46   \\
\multicolumn{1}{c|}{ABS}             & 514.15   & 514.15   & 882.58   \\
\multicolumn{1}{c|}{Rubber}          & 570.81   & 407.00   & 570.81   \\
\multicolumn{1}{c|}{MDF}             & 544.63   & 618.47   & 733.93   \\
\multicolumn{1}{c|}{Acrylic}         & 635.49   & 380.35   & 702.15   \\
\multicolumn{1}{c|}{Porcelain}       & 1276.59  & 1162.69  & 1334.07  \\
\multicolumn{1}{c|}{Glass}           & 1433.31  & 1433.31  & 1560.39  \\
\multicolumn{1}{c|}{Granite}         & 2749.87  & 2252.32  & 2749.87  \\
\multicolumn{1}{c|}{Stainless Steel} & 10184.17 & 6388.35  & 10184.17 \\
\multicolumn{1}{c|}{Aluminum}        & 17530.03 & 12767.69 & 25972.02 \\
\multicolumn{1}{c|}{Copper}          & 23049.18 & 23049.18 & 36761.16 \\ \bottomrule
\end{tabular}%
\end{center}
\vspace{-4.9mm}
\end{table}



\section*{Acknowledgment}
This work was supported in part by NSF Awards EFRI- 1137229 and IIS-1150157, the National Institute on Disability, Independent Living, and Rehabilitation Research (NIDILRR) grant 90RE5016-01-00 via RERC TechSAge, and a Google Faculty Research Award.


\bibliographystyle{IEEEtran}
\bibliography{thermal}

\end{document}